\documentclass[11pt]{article}

\usepackage[utf8]{inputenc}
\usepackage[english]{babel}
\usepackage[a4paper,margin=1in]{geometry}
\usepackage{graphicx}
\usepackage{booktabs}
\usepackage{array}
\usepackage{ragged2e}
\usepackage{caption}
\usepackage{amsmath,amssymb}
\usepackage{url}
\usepackage{enumitem}
\usepackage[round,authoryear]{natbib}
\usepackage[colorlinks=true,allcolors=blue]{hyperref}
\usepackage{longtable}

\newcommand{\keywords}[1]{\par\smallskip\noindent\textbf{Keywords:} #1}

\title{\textbf{Quantifying Emotional Tone in Tolkien's \emph{The Hobbit}:\\
Dialogue Sentiment Analysis with RegEx, NRC-VAD, and Python}}

\author{
  Lilin Qiu\\
  Department of Language Technologies and Digital Humanities\\
  University of Turin\\
  \texttt{lilin.qiu@edu.unito.it}
}

\date{}

\begin{document}

\maketitle

\begin{abstract}
This study analyzes the emotional tone of dialogue in J.\ R.\ R.\ Tolkien's \emph{The Hobbit} (1937) using computational text analysis. Dialogue was extracted with regular expressions, then preprocessed, and scored using the NRC-VAD lexicon to quantify emotional dimensions. The results show that the dialogue maintains a generally positive (high valence) and calm (low arousal) tone, with a gradually increasing sense of agency (dominance) as the story progresses. These patterns reflect the novel's emotional rhythm: moments of danger and excitement are regularly balanced by humor, camaraderie, and relief. Visualizations---including emotional trajectory graphs and word clouds---highlight how Tolkien's language cycles between tension and comfort. By combining computational tools with literary interpretation, this study demonstrates how digital methods can uncover subtle emotional structures in literature, revealing the steady rhythm and emotional modulation that shape the storytelling in \emph{The Hobbit}.

\keywords{digital philology, Tolkien, \emph{The Hobbit}, sentiment analysis, NRC-VAD, RegEx, dialogue analysis}
\end{abstract}

% ====================================================
\section{Introduction}

Philology has long been described as ``the discipline of making sense of text'' \citep{pollock2014}. It involves reading deeply to uncover meaning within language and time. Digital philology continues this mission through computational tools and open, machine-actionable workflows that allow texts to be searched, annotated, compared, and visualized dynamically \citep{crane2009}.

In this spirit, \emph{The Hobbit} by J.\ R.\ R.\ Tolkien (1937) is an ideal subject for this digital philology study. As a trained philologist, Tolkien placed language at the heart of his creative process. He once wrote, ``I am a philologist and all my work is philological'' \citep{tolkien1981}. For Tolkien, invention began with language itself rather than with plot or characters: ``The invention of languages is the foundation. The `stories' were made rather to provide a world for the languages than the reverse. To me a name comes first and the story follows'' \citep{tolkien1981}.

Scholars such as \citet{kullmann2009} have likewise argued that Tolkien's fiction is less about elves, dwarves, and warriors than about the production of meaning through linguistic conventions. This study adopts a computational perspective to explore the emotional structure of \emph{The Hobbit}, investigating how Tolkien's language encodes emotional tone and how these patterns evolve across the novel's dialogue. In keeping with Tolkien's view of language as a living, evolving whole, this study focuses on broader emotional and linguistic trends rather than isolated sentences.

The study traces the development of dialogue sentiment throughout \emph{The Hobbit}. Methodologically, it employs regular expressions (RegEx) to extract all dialogue, applies preprocessing to clean and normalize the text, and then uses the NRC-VAD Lexicon to score each chapter's valence (positive vs.\ negative), arousal (excited vs.\ calm), and dominance (in control vs.\ powerless). The results include emotional trajectory graphs and word clouds generated with Python libraries. This workflow follows the sentiment analysis pipeline outlined by \citet{hankar2025}: collect text $\rightarrow$ clean $\rightarrow$ extract features $\rightarrow$ analyze $\rightarrow$ visualize.

The goal is to transform Tolkien's crafted language into visible patterns of feeling and to relate those patterns back to the narrative. Ultimately, this work demonstrates how computation can extend the traditional philological quest to ``make sense of texts''---a digital companion to close reading.

% ====================================================
\section{Methodology}

\subsection{Text Sourcing}

The corpus for this study consists of the full text of \emph{The Hobbit} by J.\ R.\ R.\ Tolkien (1937). A plain text (.txt) version was used to ensure compatibility with the text-processing methods used later, which work best with clean, unformatted input.

\subsection{Chapter Extraction}

Regular expressions (RegEx) are a formal syntax for describing text patterns. They are commonly used for text extraction, validation, and cleaning, as well as for identifying or transforming recurring linguistic structures.

In this study, RegEx is used to extract each chapter and save it as an individual .txt file inside the \texttt{chapters} directory. Specifically, Python's regular expression (\texttt{re}) module is used for pattern matching, which allows more granular extraction. Chapters are identified by detecting text that starts with the word ``chapter'', followed by a space and then one or more digits (e.g., \emph{Chapter 1}, \emph{Chapter 2}).

\subsection{Dialogue Extraction}

Following \citet{vishnubhotla2024}, who demonstrate that narration and dialogue ``largely express disparate emotions'' with arc correlations ``close to 0'', this study focuses exclusively on dialogue to capture emotional states expressed directly in speech.

All spoken dialogue is extracted chapter by chapter using Python's \texttt{re} module and stored in .csv format inside the \texttt{dialogues} directory. Dialogue is identified by detecting text enclosed in double quotation marks. 

\subsection{Pre-processing}

As \citet{hankar2025} note, ``preprocessing collected texts is a necessary stage before performing any sort of analysis, since the quality of data matters and can directly impact further tasks.'' Accordingly, a series of pre-processing steps is applied to the dialogue files generated from the above steps. The goal is to provide clean input data for sentiment analysis that will be performed in later stages.

These steps include tokenization, normalization, contraction handling, punctuation removal, and stopword removal. For simplicity, all these processes were implemented within the same script.

\subsubsection{Tokenization, Normalization, Contraction Handling, and Punctuation Removal}

Normalization standardizes the text by converting all characters to lowercase and removing extraneous spaces. This ensures that words such as ``Happy'', ``happy'', and ``HAPPY'' are treated as identical. Tokenization then divides the text into smaller word units (tokens), allowing each word to be processed and analyzed separately.

In addition, contraction handling is included to ensure that shortened negative forms such as ``don't'' and ``can't'' are mapped to a representation that explicitly preserves negation. In the preprocessing pipeline, auxiliary--negation constructions (e.g., ``do~+~n't'', ``ca~+~n't'', ``wo~+~n't'') are replaced with the token ``not'' so that the presence of negation is made explicit rather than being hidden inside a contracted form. Making negation overt in this way prevents it from being lost during tokenization and ensures that later sentiment computations are based on an accurate representation of the original text. 

Finally, punctuation removal is applied to simplify the dataset and focus only on meaningful lexical items that contribute to sentiment. Since punctuation marks do not carry intrinsic emotional value, removing them helps to create cleaner word tokens that can be directly matched with entries in the NRC-VAD lexicon for valence, arousal, and dominance scoring.

Together, these preprocessing steps standardize the textual data, retain key sentiment-bearing features such as negations, and reduce noise in preparation for lexicon-based sentiment analysis \citep{hankar2025}.

\subsubsection{Stopword Removal}

Stopwords are words such as \emph{the}, \emph{and}, \emph{of}, \emph{to}, and \emph{at} that appear frequently in almost all texts but do not carry emotional or semantic meaning that helps detect sentiment. Removing stopwords therefore makes the data cleaner and more focused on meaningful words, while also reducing computational load and making sentiment analysis faster and more efficient \citep{hankar2025}.

The NLTK stopword list was found to be insufficient, as in previous analyses words like ``would'', ``could'', ``may'' and other common words still appeared frequently in the word clouds, creating noise that complicates interpretation. To address this, an extended list of English stopwords from a public GitHub Gist\footnote{\url{https://gist.githubusercontent.com/rishg2/35e00abf8941d72d419224cfd5b5925d/raw/12d899b70156fd0041fa9778d657330b024b959c/stopwords.txt}} is integrated into the script.

\begin{figure}[htbp]
    \centering
    \includegraphics[width=0.7\textwidth]{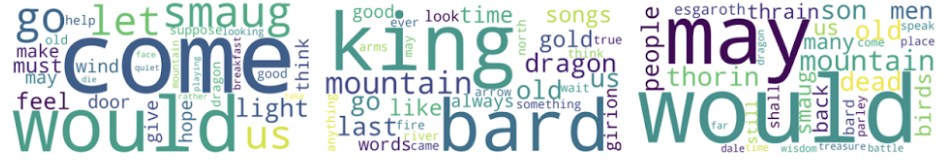}
    \caption{Word cloud before stopword removal. High-frequency, non-emotional words such as ``would'', ``may'', and ``come'' remain prominent, clouding interpretation.}
    \label{fig:wordcloud-before}
\end{figure}

Using the stopwords removal script, which also integrates tokenization, normalization, contraction handling, and punctuation removal, it is possible to remove common stopwords such as ``the'' and ``and'', as well as additional high-frequency, non-emotional words such as ``would'', ``could'' and ``come'' with the extended stopword list. The cleaned and stopword-filtered dialogues are then saved into the \texttt{dialogues\_filtered} folder for subsequent sentiment analysis.

\subsection{Compilation}

All chapter-level dialogue files are concatenated into a single text file representing the complete set of spoken tokens in \emph{The Hobbit}. This allows both chapter-specific and overall sentiment analyses.

Specifically, all the \texttt{chapter\_x\_dialogues.csv} files are concatenated into a single .txt file named \texttt{full\_dialogue.txt} for later use.

The cleaned .txt document is then uploaded to Voyant Tools' Cirrus for word frequency visualization. The Cirrus output shows \emph{good} (89), \emph{time} (65), \emph{Baggins} (46), \emph{mountain} (42), and \emph{Thorin} (41) as the most frequently appearing dialogue words. These prominent terms suggest a tone that is reflective, relational, and goal-oriented. Words such as \emph{good} and \emph{time} evoke a sense of warmth, reassurance, and social connection within the characters' interactions, while \emph{Baggins} and \emph{Thorin} highlight the centrality of personal identity and leadership in the dialogue. The frequent mention of \emph{mountain} in the characters' speech underscores how the quest and the journey toward it remain central preoccupations throughout their conversations. Together, these dominant words convey a mood that is hopeful yet purposeful, capturing the blend of companionship and determination that characterizes \emph{The Hobbit}'s storytelling.

\begin{figure}[htbp]
    \centering
    \includegraphics[width=0.7\textwidth]{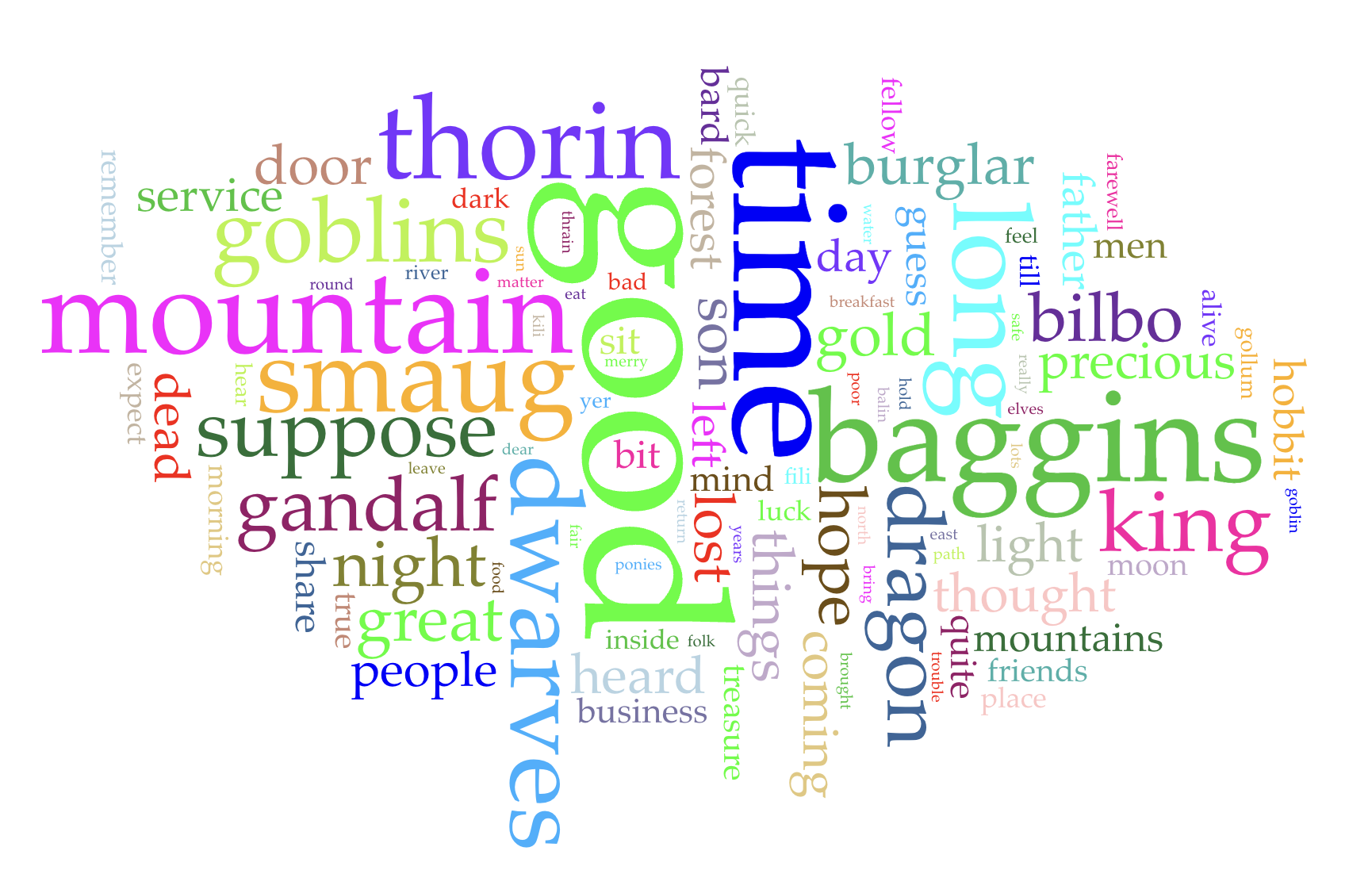}
    \caption{Cirrus visualization of high-frequency words from the full dialogue in \emph{The Hobbit}, generated using Voyant Tools.}
    \label{fig:cirrus}
\end{figure}

% ====================================================
\subsection{Sentiment Analysis and Visualisation}

For this stage, the NRC-VAD Lexicon is used to analyze the emotional tone of the dialogues. This lexicon assigns each English word three psychological scores---Valence, Arousal, and Dominance (VAD)---each ranging from 0 (low) to 1 (high). As described in lexicon-based approaches, ``lexicons are essentially dictionaries specifically designed for sentiment analysis \dots each token is associated with a predefined sentiment score indicating its intensity'' \citep{hankar2025}.

Within this lexicon-based framework:
\begin{itemize}[noitemsep]
    \item Valence shows how positive or negative a word is (e.g., \emph{joy} = high valence, \emph{death} = low valence).
    \item Arousal measures the level of emotional excitement (e.g., \emph{terror} = high; \emph{relaxed} = low).
    \item Dominance describes how much control or power a word expresses (e.g., \emph{leader} = high, \emph{victim} = low).
\end{itemize}

The NRC-VAD Lexicon, developed by \citet{mohammad2018vad, mohammad2025vad}, includes more than 55{,}000 English words and phrases rated by human annotators on these three emotional dimensions. According to \citet{mohammad2025vad}, ``the three primary independent dimensions of emotions are valence or pleasure (positiveness–negativeness/pleasure–
displeasure), arousal (active–passive), and dominance (dominant–submissive).'' The lexicon was built through large-scale human annotation and achieved very high reliability. As Mohammad notes, ``The large number of entries in the VAD Lexicon and the high reliability of the scores make it useful for a number of research projects and applications.''

Because it is domain-independent and psychologically interpretable, the NRC-VAD Lexicon is well suited for exploring how emotion shifts across the progression of dialogue in \emph{The Hobbit}. By mapping each word in the dialogue to its VAD scores, it is possible to model changes in valence (positivity), arousal (emotional intensity), and dominance (sense of agency) across the chapters.

For data processing, NumPy is used for numerical calculation and Pandas for handling the results in table form. The average VAD scores are calculated for each chapter and then visualised using Matplotlib and Seaborn.

The resulting line charts show how valence, arousal, and dominance rise and fall throughout \emph{The Hobbit}, giving a visual sense of the emotional rhythm within the dialogue. Word clouds are generated using Python's \texttt{wordcloud} library to highlight the most frequent words in each chapter. This helps cross-reference the lexical and emotional data---showing, for example, how certain recurring words might coincide with emotional highs or lows. While Voyant Tools' Cirrus was used earlier for general word frequency visualisation, the \texttt{wordcloud} library is preferred here for its flexibility and seamless integration with the Python workflow.

As \citet{hankar2025} point out, visualisations such as ``bar charts, pie charts, word clouds'' are important for communicating sentiment results clearly. In this project, the combination of VAD line graphs and word clouds provides both a broad emotional overview and a close look at the language patterns that shape the tone of the dialogues in \emph{The Hobbit}.

\begin{figure}[htbp]
    \centering
    \includegraphics[width=0.75\textwidth]{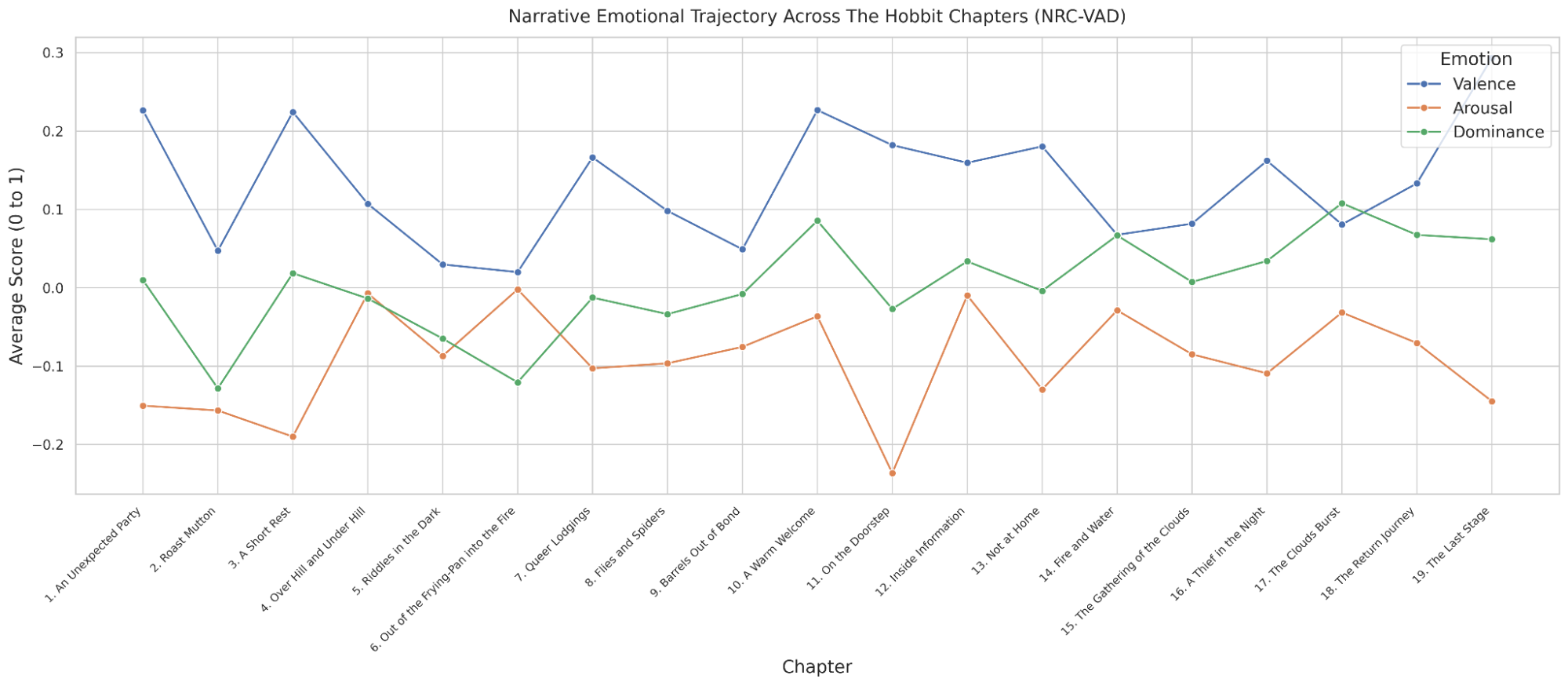}
    \caption{Emotional trajectory of valence, arousal, and dominance across the nineteen chapters of \emph{The Hobbit}, based on NRC-VAD scores for filtered dialogue.}
    \label{fig:trajectory-filtered}
\end{figure}

\begin{figure}[htbp]
    \centering
    \includegraphics[width=\textwidth]{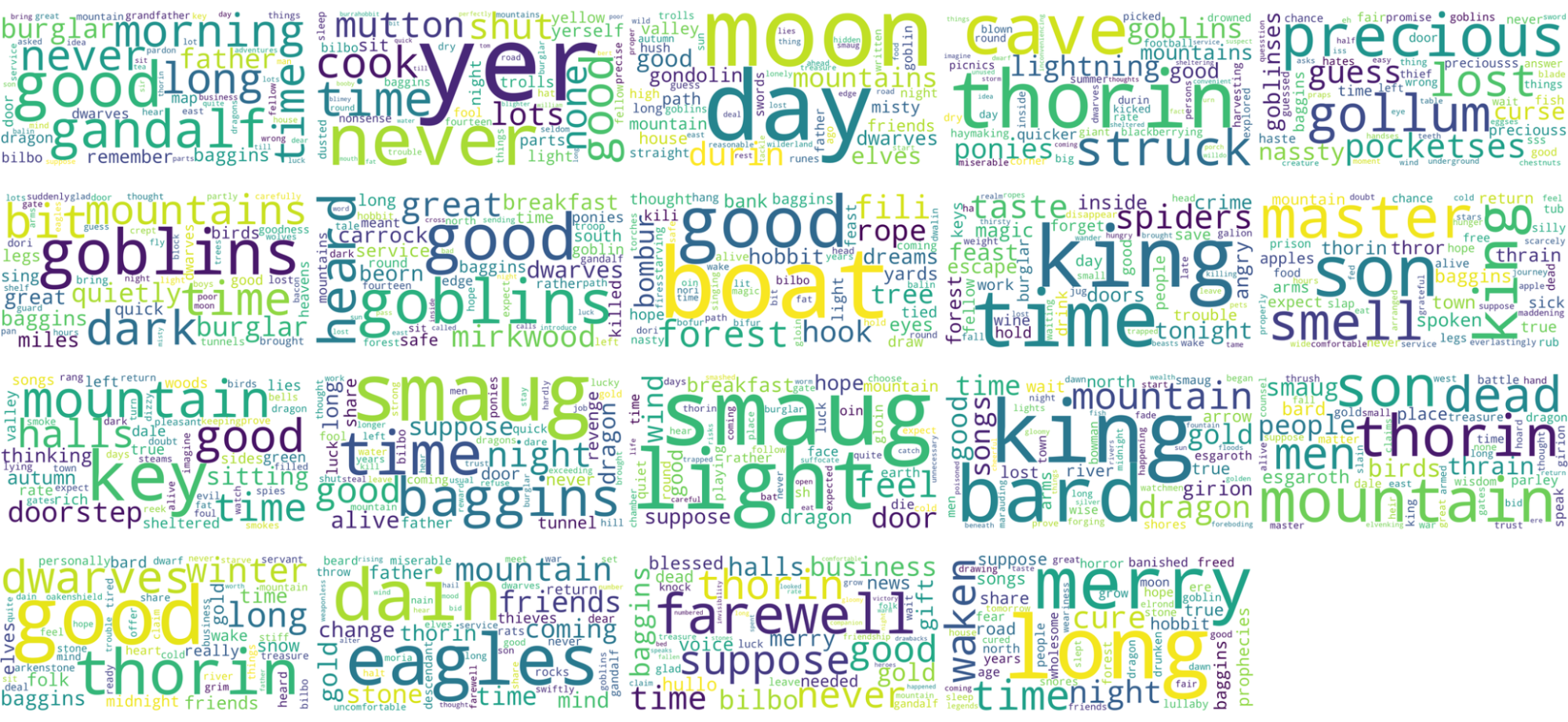}
    \caption{Word cloud grid of all 19 chapters of \emph{The Hobbit}, generated from high-frequency words in each chapter's filtered dialogue.}
    \label{fig:wordcloud-grid-filtered}
\end{figure}

% ====================================================
\section{Findings}

\subsection{Emotional Peaks and Lows}

The VAD sentiment graph (Figure~\ref{fig:trajectory-filtered}) highlights prominent peaks and troughs:

\medskip
\noindent\textbf{High Valence:} Chapter 3 (``A Short Rest''), Chapter 10 (``A Warm Welcome''), Chapter 19 (``The Last Stage'').

\noindent\textbf{Low Valence:} Chapter 5 (``Riddles in the Dark''), Chapter 6 (``Out of the Frying-Pan into the Fire''), Chapter 9 (``Barrels Out of Bond'').

\noindent\textbf{High Arousal:} Chapter 4 (``Over Hill and Under Hill''), Chapter 6 (``Out of the Frying-Pan into the Fire''), Chapter 12 (``Inside Information'').

\noindent\textbf{Low Arousal:} Chapter 2 (``Roast Mutton''), Chapter 3 (``A Short Rest''), Chapter 11 (``On the Doorstep'').

\noindent\textbf{High Dominance:} Chapter 10 (``A Warm Welcome''), Chapter 14 (``Fire and Water''), Chapter 17 (``The Clouds Burst'').

\noindent\textbf{Low Dominance:} Chapter 2 (``Roast Mutton''), Chapter 5 (``Riddles in the Dark''), Chapter 6 (``Out of the Frying-Pan into the Fire'').

In Tolkien's dialogue, emotion is inseparable from language: how characters speak reflects how they feel and how they move as a group. The following table cross-references the NRC-VAD results with each chapter's most frequent dialogue words. For each selected chapter, plot summary and lexical interpretation are provided to illustrate how specific word choices reflect emotional tone within narrative context.

{%
\footnotesize
\setlength{\tabcolsep}{3pt}
\renewcommand{\arraystretch}{1.15}

\begin{longtable}{
  >{\RaggedRight\arraybackslash}p{0.11\textwidth}  % Chapter (narrower)
  >{\RaggedRight\arraybackslash}p{0.20\textwidth}  % 10 words (narrower)
  >{\centering\arraybackslash}p{0.32\textwidth}    % Word cloud (wider)
  >{\RaggedRight\arraybackslash}p{0.37\textwidth}  % Interpretation (wider)
}
\toprule
\textbf{Chapter (Title)} & \textbf{10 Most Frequent Words} & \textbf{Word Cloud} & \textbf{Plot and Lexical Interpretation} \\
\midrule
\endfirsthead

\toprule
\textbf{Chapter (Title)} & \textbf{10 Most Frequent Words} & \textbf{Word Cloud} & \textbf{Plot and Lexical Interpretation} \\
\midrule
\endhead

\bottomrule
\endfoot

% ============== HIGH VALENCE ==============
\multicolumn{4}{l}{\textbf{High Valence (highly positive)}} \\
\midrule

\vspace{0pt}Ch.\ 3 -- \emph{A Short Rest} &
\vspace{0pt}\emph{moon, day, durin, mountains, elves, good, gondolin, dwarves, mountain, misty} &
\vspace{0pt}\includegraphics[width=0.95\linewidth]{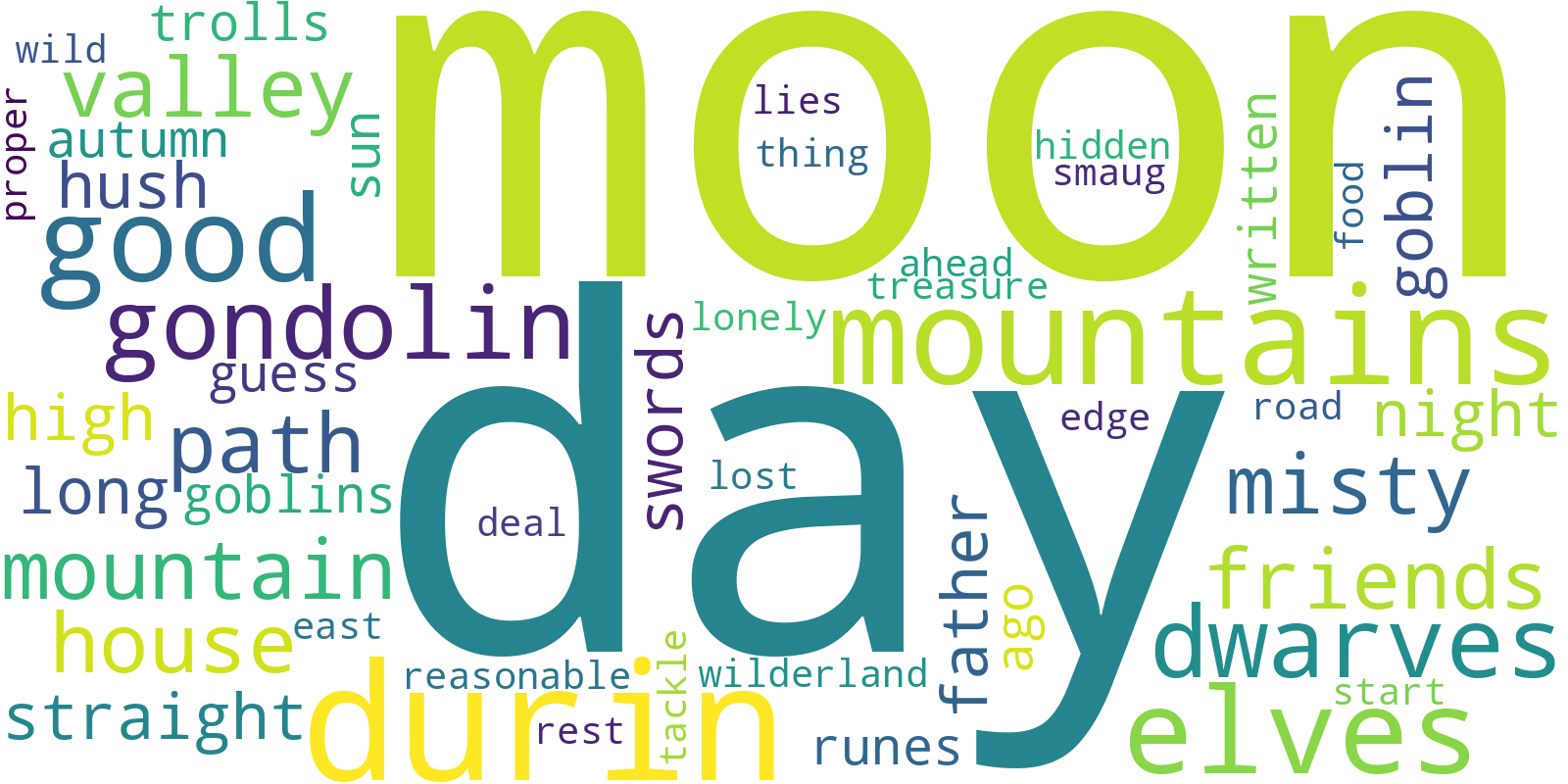} &
\vspace{0pt}The company’s stop at Rivendell marks the first respite in their arduous journey. The elves’ songs, Elrond’s hospitality, and the natural beauty of the valley create an atmosphere of calm and renewal. Dialogue is lyrical and descriptive, full of references to moonlight, mountains, and dwarves’ ancestral heritage, and frequent use of words like \emph{good} and \emph{elves} conveys gratitude and unity. \\[0.4em]

\vspace{0pt}Ch.\ 10 -- \emph{A Warm Welcome} &
\vspace{0pt}\emph{son, king, master, smell, baggins, thorin, thrain, thror, town, spoken} &
\vspace{0pt}\includegraphics[width=0.95\linewidth]{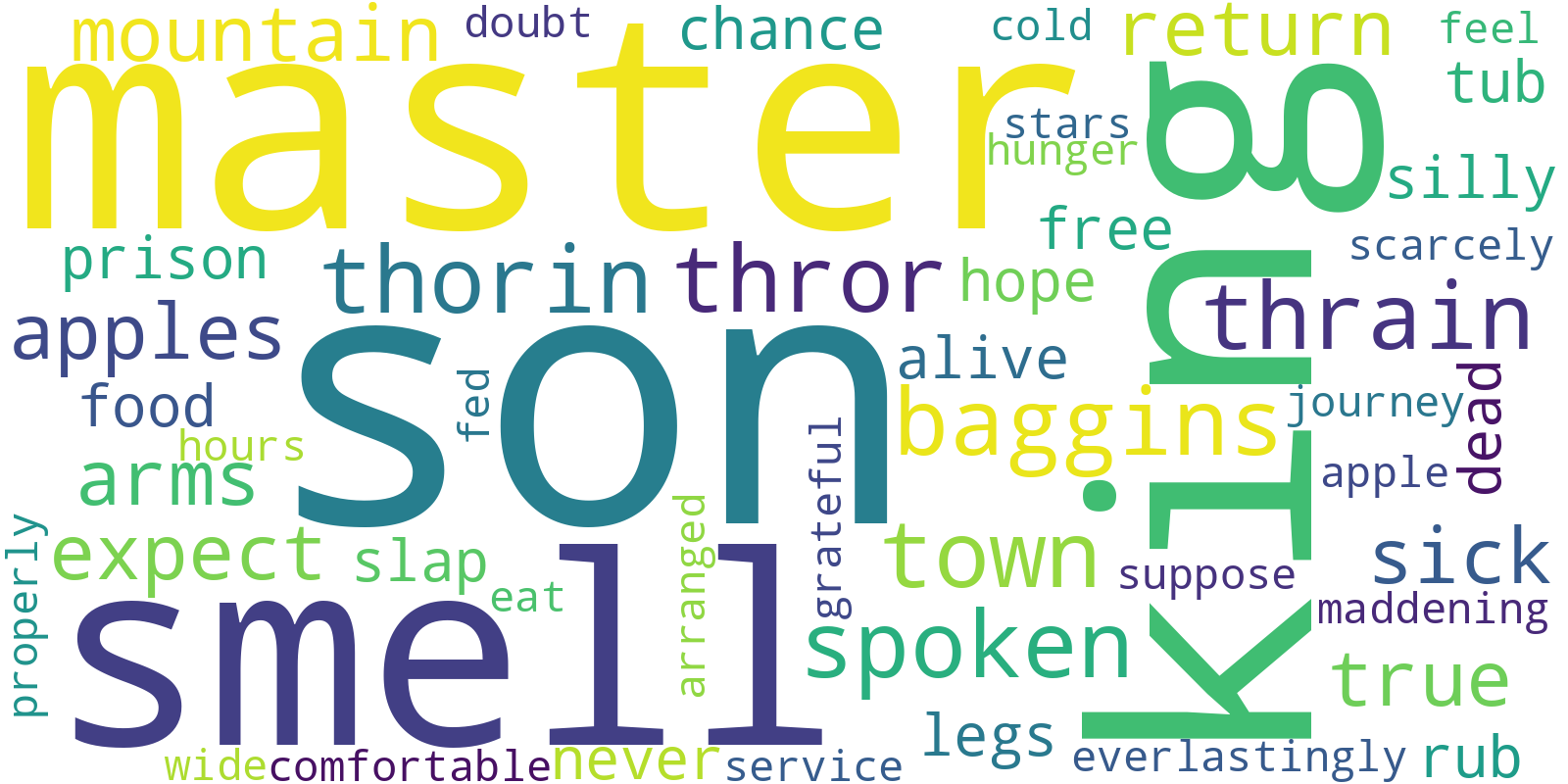} &
\vspace{0pt}Arriving at Lake-town, Thorin and his company are hailed as heroes destined to reclaim the Lonely Mountain. Dialogue becomes ceremonial and confident, filled with titles and names—\emph{king}, \emph{master}, \emph{Thorin}, \emph{Thrain}, \emph{Thror}—that evoke lineage and legitimacy. The tone of the exchanges radiates optimism and collective pride. \\[0.4em]

\vspace{0pt}Ch.\ 19 -- \emph{The Last Stage} &
\vspace{0pt}\emph{long, merry, time, waken, night, cure, suppose, road, share, hobbit} &
\vspace{0pt}\includegraphics[width=0.95\linewidth]{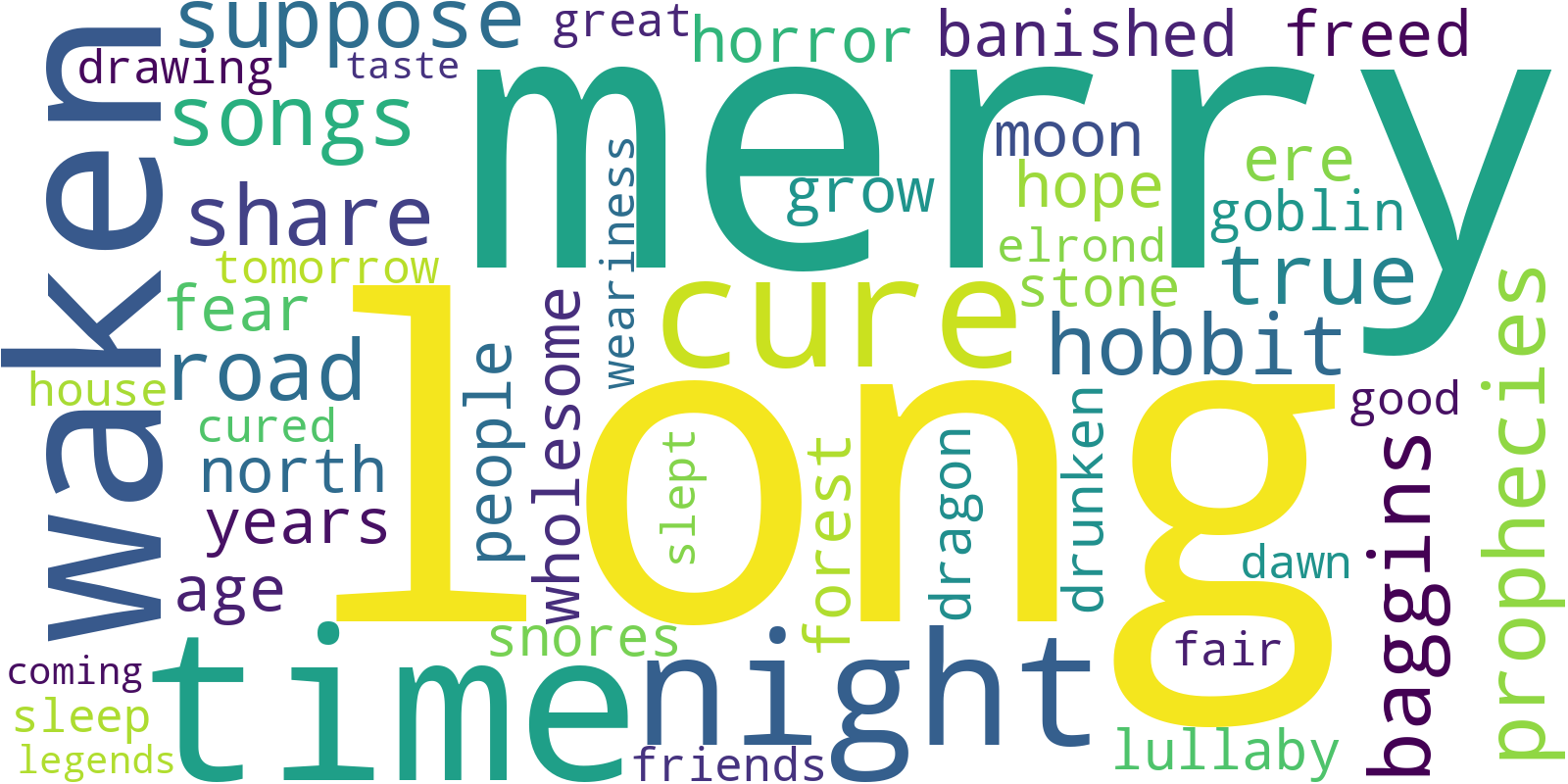} &
\vspace{0pt}The closing chapter completes Bilbo’s emotional arc. Returning home after the adventure, he speaks with affection and gentle humor, reflecting on \emph{time}, \emph{road}, and \emph{hobbit} life. Soft, domestic language and words like \emph{merry}, \emph{cure} and \emph{share} underscore restoration, peace, and gratitude. \\
\midrule

% ============== LOW VALENCE ==============
\multicolumn{4}{l}{\textbf{Low Valence (highly negative)}} \\
\midrule

\vspace{0pt}Ch.\ 5 -- \emph{Riddles in the Dark} &
\vspace{0pt}\emph{precious, gollum, lost, pocketses, guess, nassty, curse, goblinses, preciouss, baggins} &
\vspace{0pt}\includegraphics[width=0.95\linewidth]{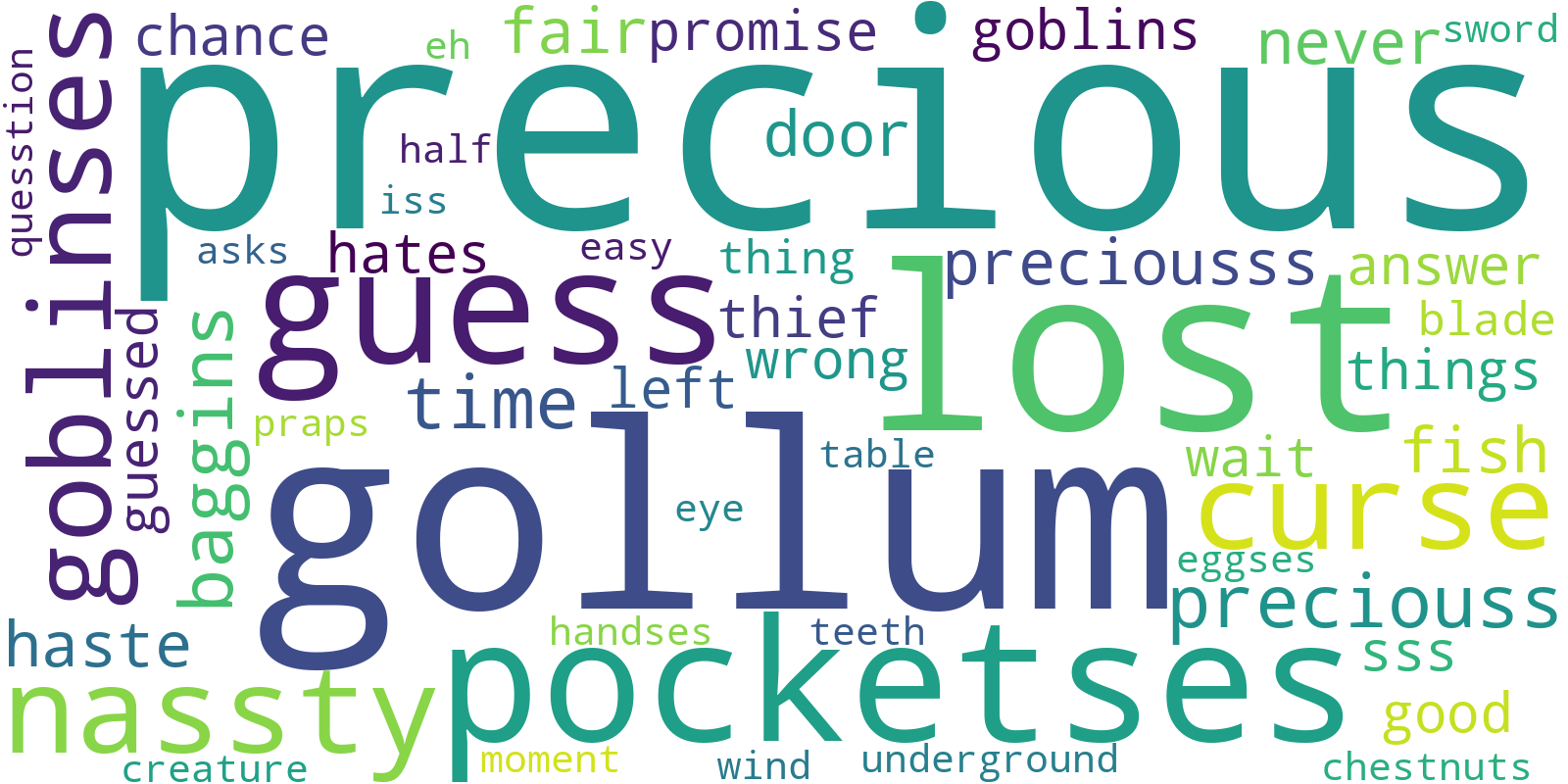} &
\vspace{0pt}This chapter represents one of the emotional nadirs of the novel. Bilbo’s encounter with Gollum in the deep caves is shrouded in isolation, fear, and ambiguity. Dialogue dominated by Gollum’s obsessive \emph{precious} and distorted forms like \emph{nassty} and \emph{goblinses} creates menace, while words such as \emph{lost} and \emph{curse} emphasise hostility and claustrophobia. \\[0.4em]

\vspace{0pt}Ch.\ 6 -- \emph{Out of the Frying-Pan into the Fire} &
\vspace{0pt}\emph{goblins, dark, time, bit, mountains, burglar, quietly, baggins, great, miles} &
\vspace{0pt}\includegraphics[width=0.95\linewidth]{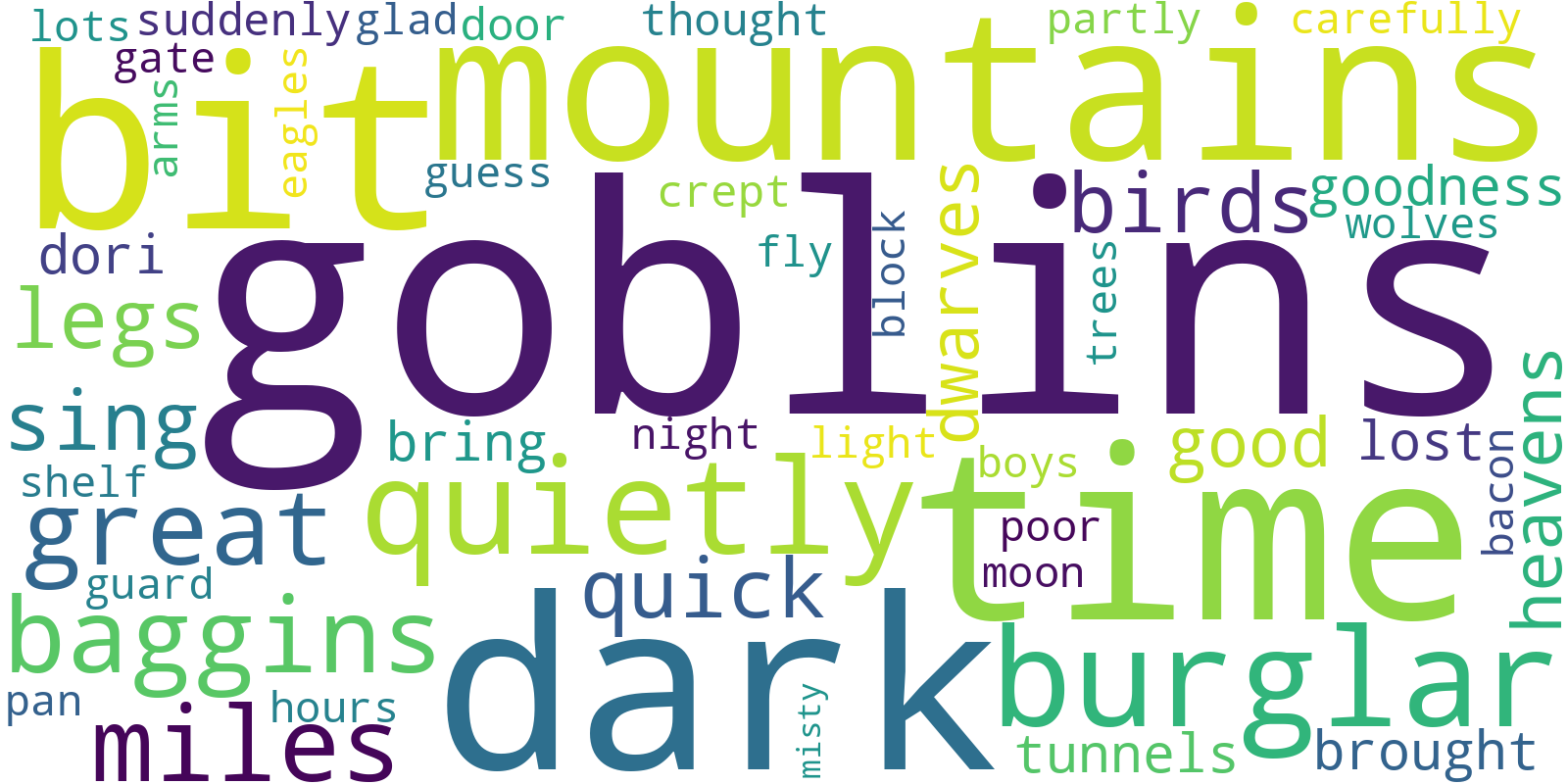} &
\vspace{0pt}The group escapes goblins only to face wolves and fire. Fragmented, breathless moments in the scene reflect panic and disorientation. Frequent references to \emph{dark}, \emph{mountains}, and \emph{goblins} stress pursuit and exhaustion; even Bilbo’s attempts at bravery are voiced through anxious, low-valence speech. \\[0.4em]

\vspace{0pt}Ch.\ 9 -- \emph{Barrels Out of Bond} &
\vspace{0pt}\emph{king, time, spiders, taste, tonight, feast, escape, magic, doors, inside} &
\vspace{0pt}\includegraphics[width=0.95\linewidth]{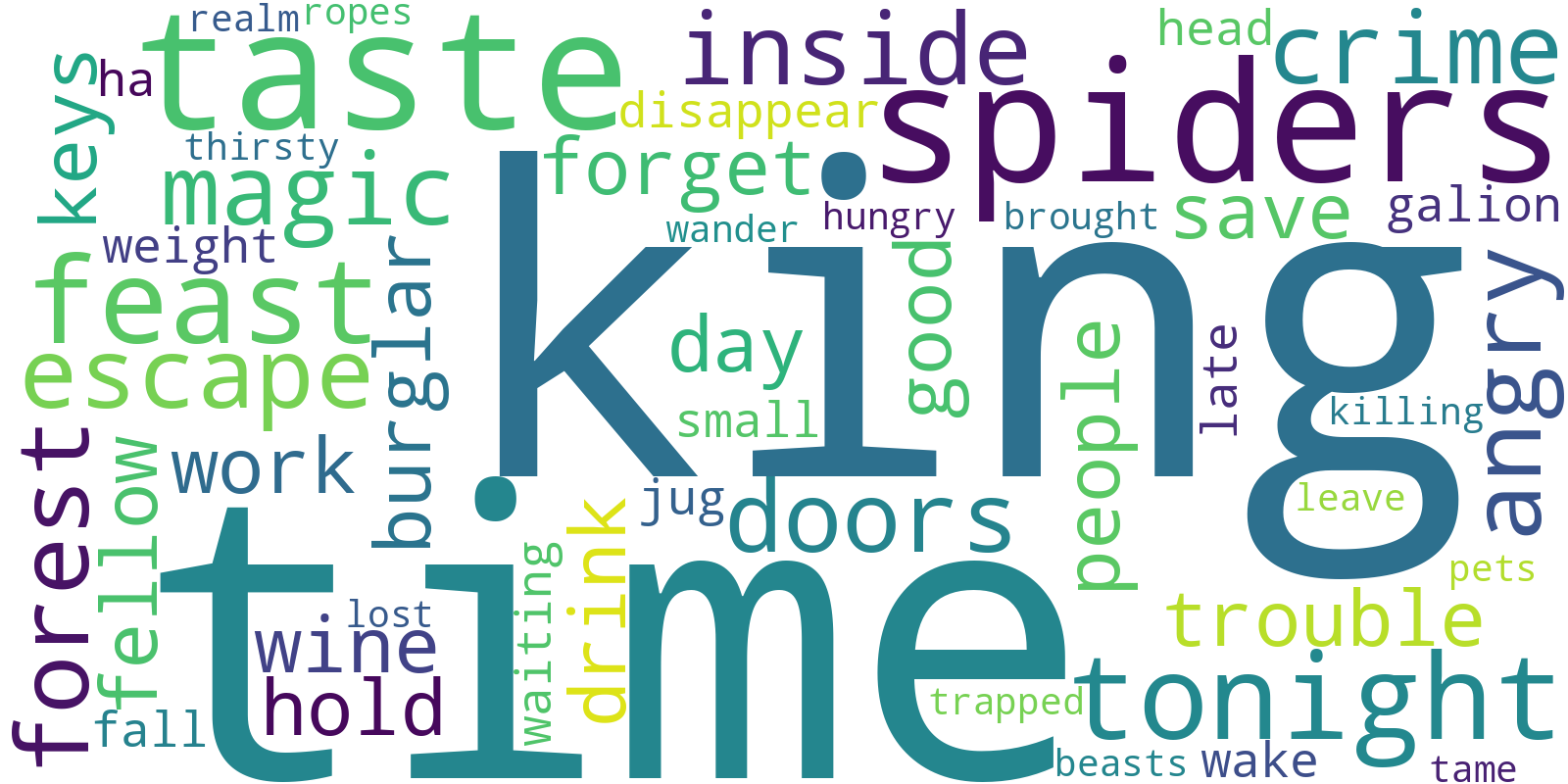} &
\vspace{0pt}Imprisoned by the Wood-Elves, the dwarves depend on Bilbo’s hidden planning. Dialogue shifts toward restraint and calculation, focusing on \emph{escape}, \emph{doors}, and \emph{inside}. Low valence reflects confinement and tension, with emerging hints of agency. \\
\midrule

% ============== HIGH AROUSAL ==============
\multicolumn{4}{l}{\textbf{High Arousal (highly excited)}} \\
\midrule

\vspace{0pt}Ch.\ 4 -- \emph{Over Hill and Under Hill} &
\vspace{0pt}\emph{thorin, cave, struck, lightning, ponies, goblins, mountains, good, quicker, summer} &
\vspace{0pt}\includegraphics[width=0.95\linewidth]{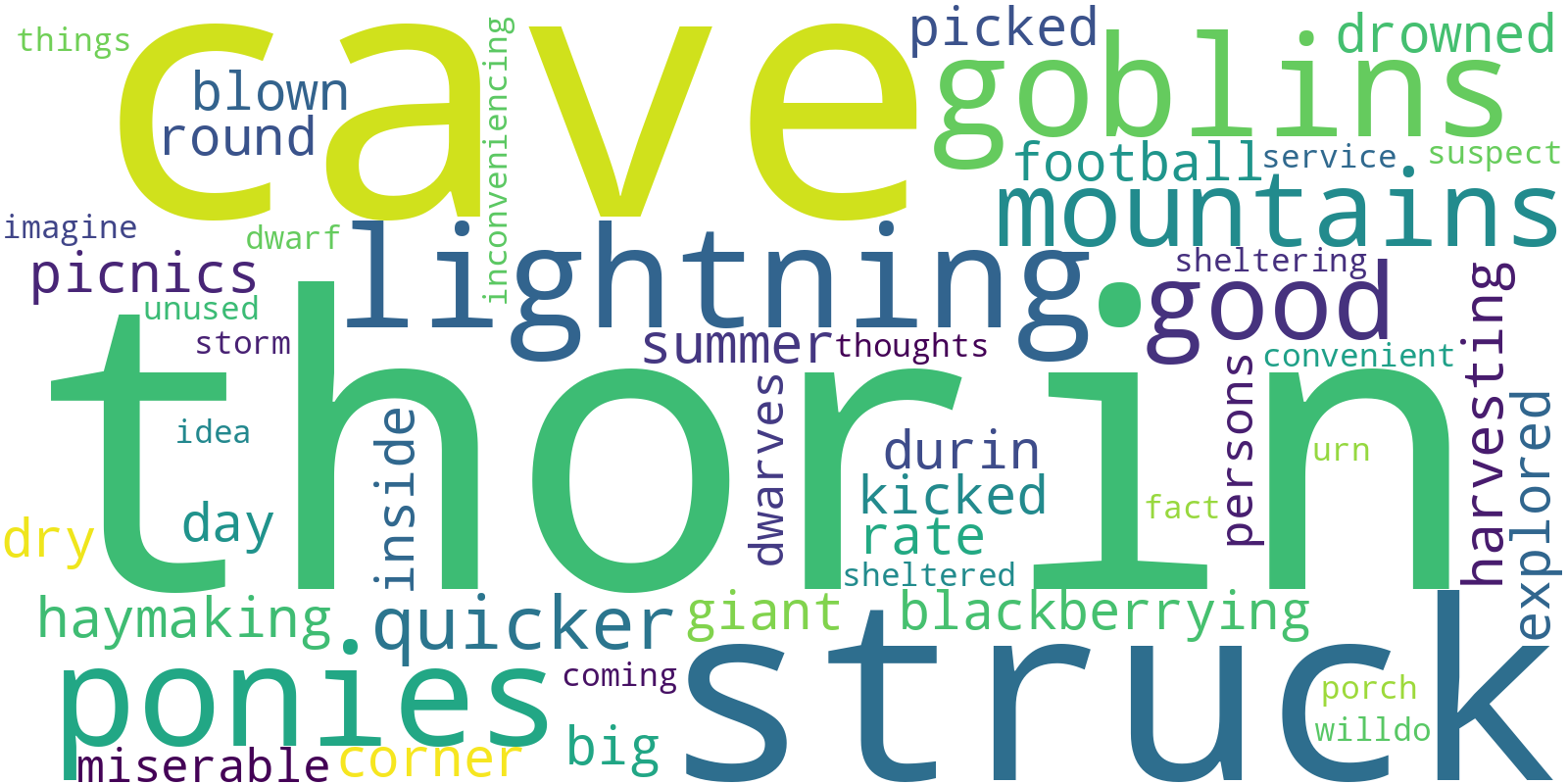} &
\vspace{0pt}A violent storm and goblin attack propel the story into action. Dialogue is clipped and sensory, with verbs like \emph{struck} and \emph{quicker} and references to \emph{lightning}, \emph{cave}, and \emph{goblins}. The language mirrors the speed and chaos of escape, creating a high-arousal spike. \\[0.4em]

\vspace{0pt}Ch.\ 6 -- \emph{Out of the Frying-Pan into the Fire} &
\vspace{0pt}\emph{goblins, dark, time, bit, mountains, burglar, quietly, baggins, great, miles} &
\vspace{0pt}\includegraphics[width=0.95\linewidth]{figures/wc_ch06_frying_pan} &
\vspace{0pt}Here the chapter epitomizes kinetic tension. Abrupt commands and nervous exclamations mix with movement-related words—\emph{quietly}, \emph{great}, \emph{miles}—to evoke restless energy. Emotionally it blends physical danger with the exhilaration of survival. \\[0.4em]

\vspace{0pt}Ch.\ 12 -- \emph{Inside Information} &
\vspace{0pt}\emph{smaug, time, baggins, good, night, suppose, dragon, alive, long, tunnel} &
\vspace{0pt}\includegraphics[width=0.95\linewidth]{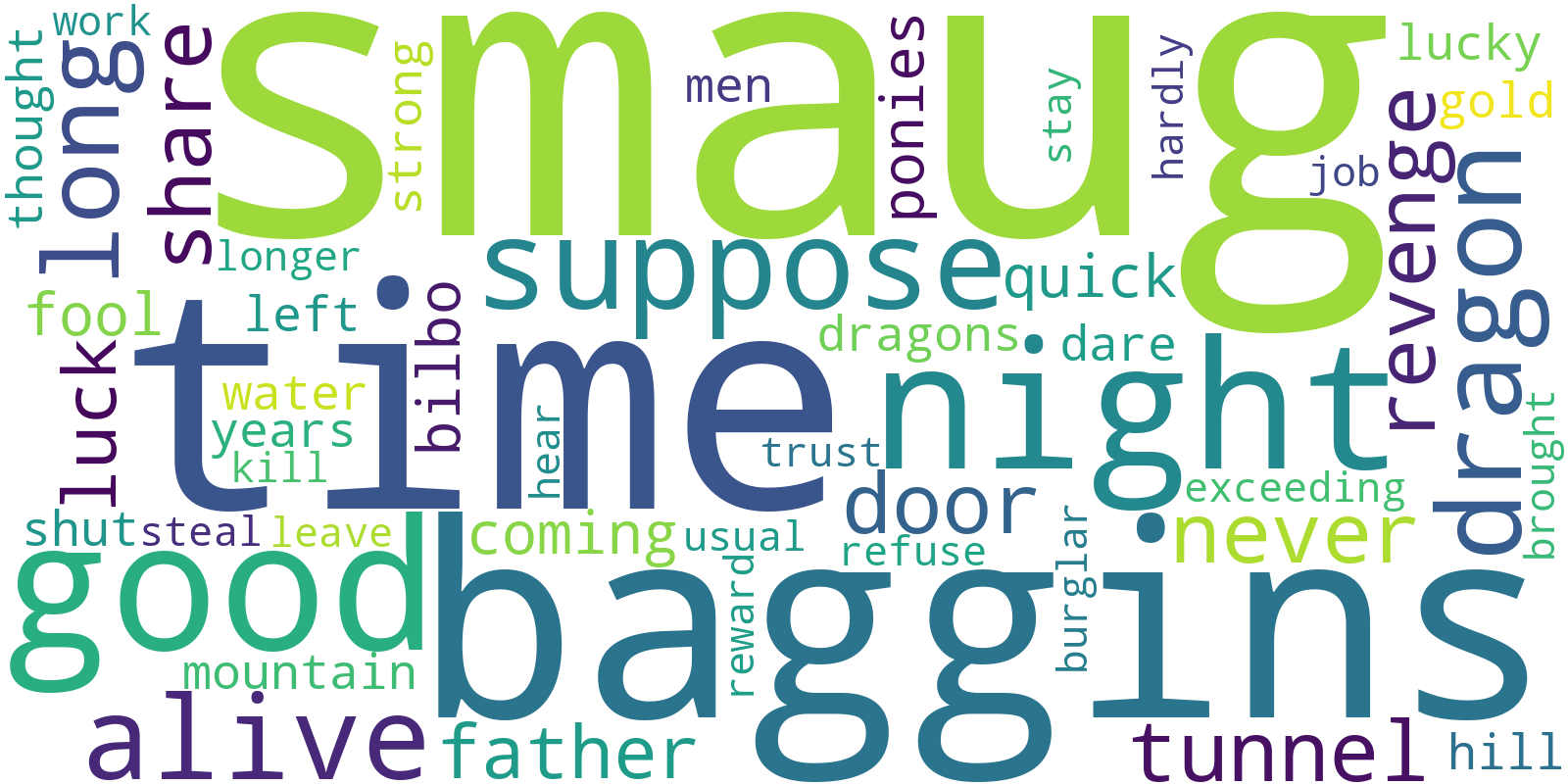} &
\vspace{0pt}Bilbo’s dialogue with Smaug balances terror and awe. Terms such as \emph{dragon}, \emph{alive}, and \emph{tunnel} heighten suspense, while careful, courteous phrasing signals self-control. High arousal is moderated by wit and diplomacy. \\
\midrule

% ============== LOW AROUSAL ==============
\multicolumn{4}{l}{\textbf{Low Arousal (highly passive)}} \\
\midrule

\vspace{0pt}Ch.\ 2 -- \emph{Roast Mutton} &
\vspace{0pt}\emph{yer, never, time, good, cook, none, mutton, shut, lots, yerself} &
\vspace{0pt}\includegraphics[width=0.95\linewidth]{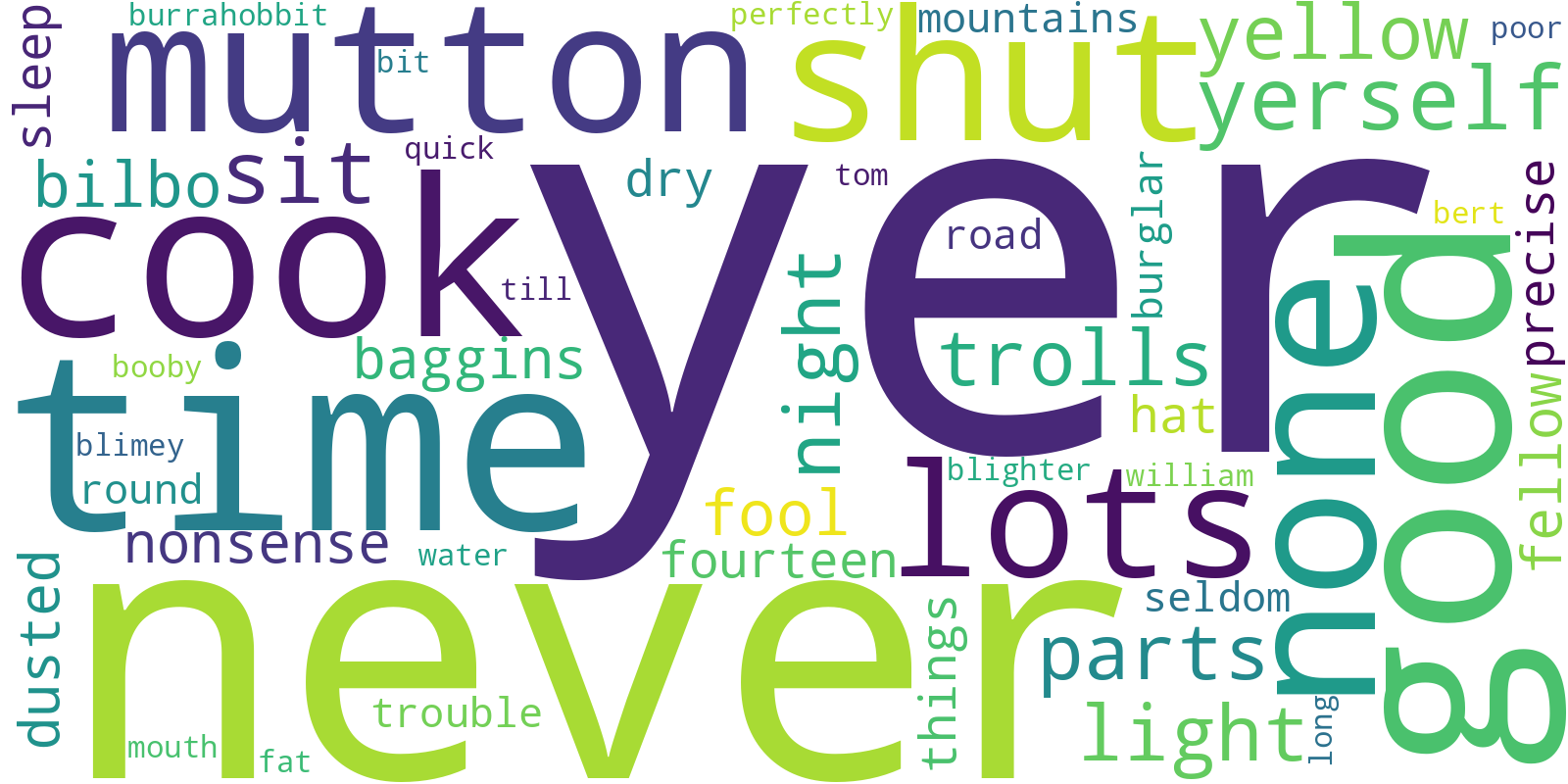} &
\vspace{0pt}The troll episode has more comic than existential danger. Dialectal forms like \emph{yer} and \emph{yerself} and food-related vocabulary (\emph{mutton}, \emph{cook}) create a mundane, domestic rhythm. Slow, conversational pacing lowers arousal even as conflict unfolds. \\[0.4em]

\vspace{0pt}Ch.\ 3 -- \emph{A Short Rest} &
\vspace{0pt}\emph{moon, day, durin, mountains, elves, good, gondolin, dwarves, mountain, misty} &
\vspace{0pt}\includegraphics[width=0.95\linewidth]{figures/wc_ch03_short_rest} &
\vspace{0pt}This Rivendell interlude sustains tranquility. Songlike exchanges and repetitions of \emph{moon}, \emph{day}, and \emph{mountain} slow the narrative tempo and reinforce rest, reflection, and gentle orientation before further trials. \\[0.4em]

\vspace{0pt}Ch.\ 11 -- \emph{On the Doorstep} &
\vspace{0pt}\emph{key, mountain, good, halls, time, doorstep, sitting, thinking, autumn, rate} &
\vspace{0pt}\includegraphics[width=0.95\linewidth]{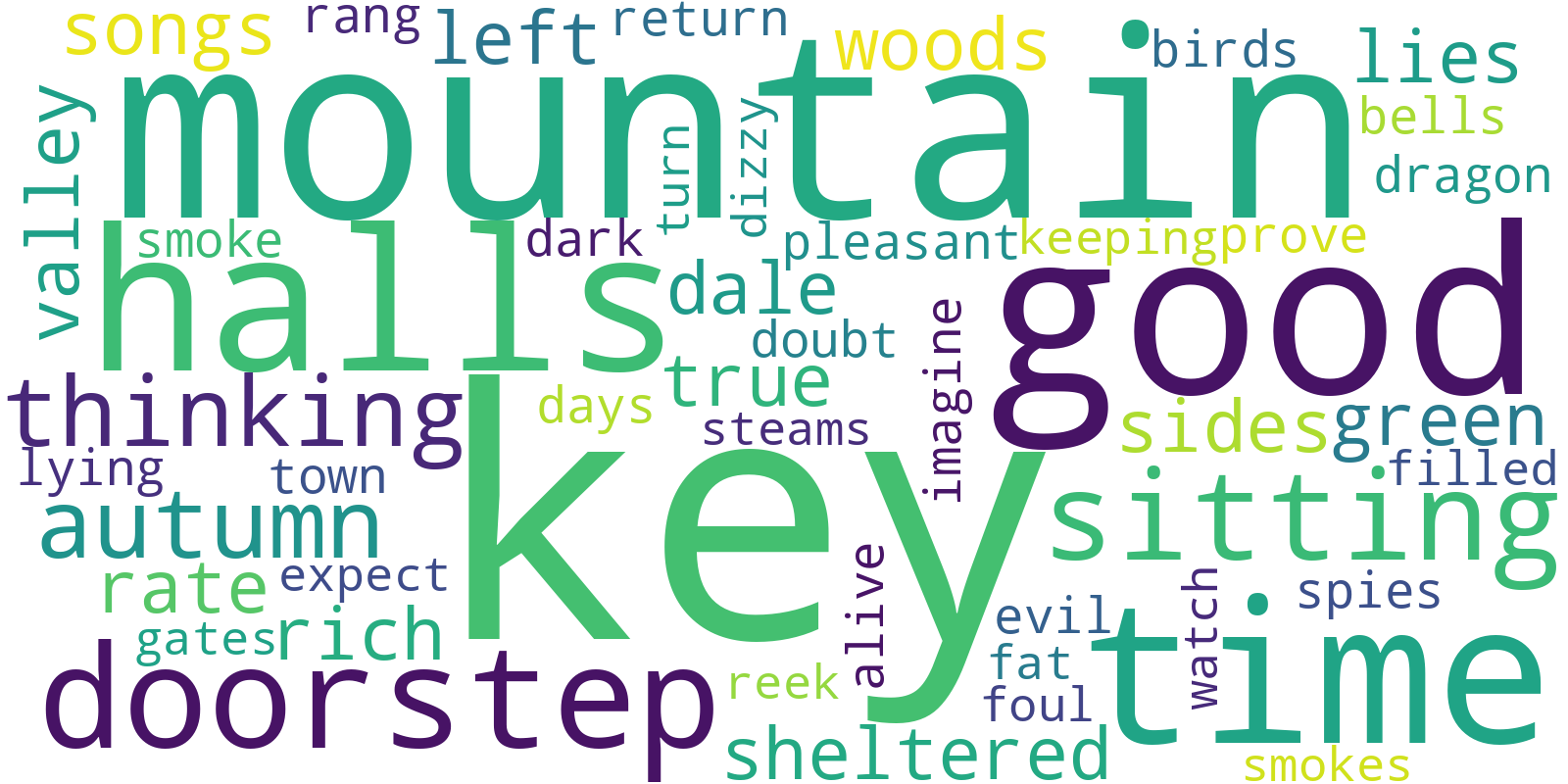} &
\vspace{0pt}Suspense arises from waiting rather than action. The company sits before Erebor’s secret door, and words like \emph{sitting}, \emph{thinking}, and \emph{autumn} capture meditative stillness and quiet anticipation. \\
\midrule

% ============== HIGH DOMINANCE ==============
\multicolumn{4}{l}{\textbf{High Dominance (highly powerful)}} \\
\midrule

\vspace{0pt}Ch.\ 10 -- \emph{A Warm Welcome} &
\vspace{0pt}\emph{son, king, master, smell, baggins, thorin, thrain, thror, town, spoken} &
\vspace{0pt}\includegraphics[width=0.95\linewidth]{figures/wc_ch10_warm_welcome} &
\vspace{0pt}Recognition and authority define this chapter. Thorin’s formal speech before the Master of Lake-town projects confidence and command, and names and titles—\emph{Thorin}, \emph{Thrain}, \emph{Thror}, \emph{king}, \emph{master}—anchor the dialogue in lineage and legitimacy. \\[0.4em]

\vspace{0pt}Ch.\ 14 -- \emph{Fire and Water} &
\vspace{0pt}\emph{king, bard, mountain, dragon, gold, time, songs, good, girion, north} &
\vspace{0pt}\includegraphics[width=0.95\linewidth]{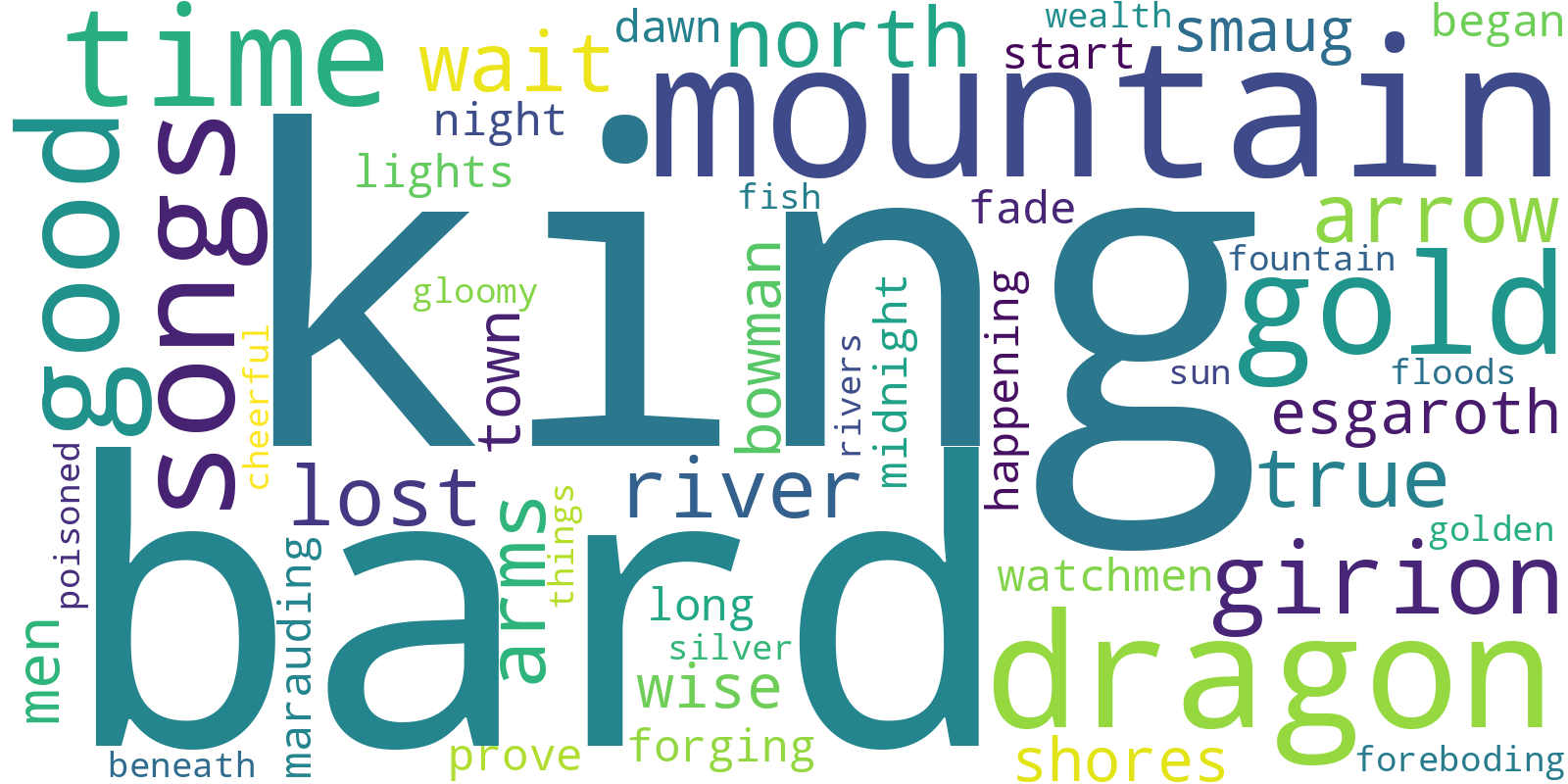} &
\vspace{0pt}Bard’s confrontation with Smaug is a decisive moment of human agency. Dialogue, though sparse, resonates with moral strength and collective resolve; \emph{king}, \emph{songs}, \emph{Girion}, and \emph{mountain} link present action to heroic legacy and courage. \\[0.4em]

\vspace{0pt}Ch.\ 17 -- \emph{The Clouds Burst} &
\vspace{0pt}\emph{dain, eagles, mountain, friends, coming, gold, stone, time, thorin, mind} &
\vspace{0pt}\includegraphics[width=0.95\linewidth]{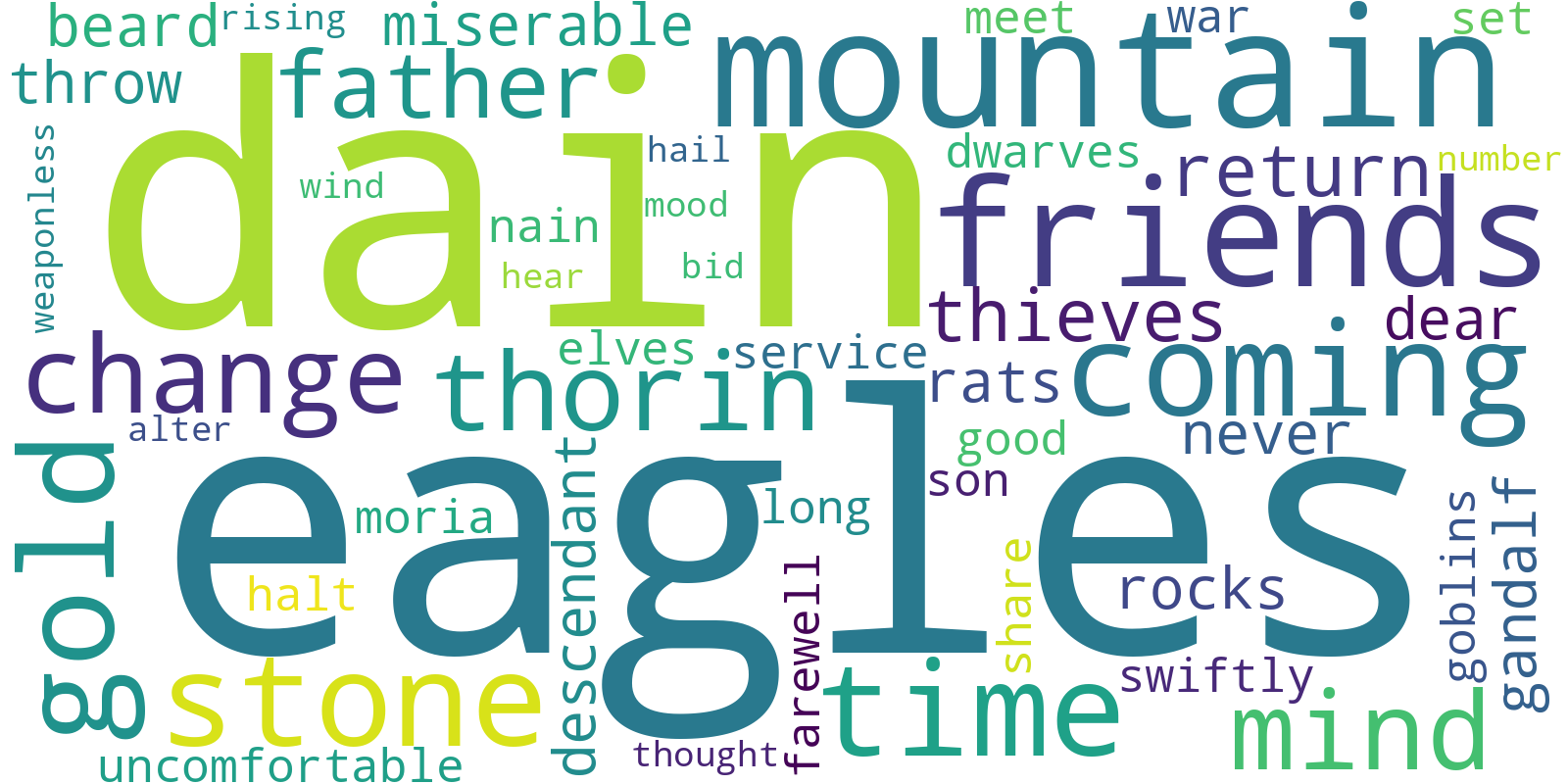} &
\vspace{0pt}Before and during the Battle of Five Armies, dialogue stresses coordination and discipline. Words such as \emph{friends}, \emph{coming}, and \emph{mind} signal unity, control, and composed readiness rather than panic, expressing high dominance. \\
\midrule

% ============== LOW DOMINANCE ==============
\multicolumn{4}{l}{\textbf{Low Dominance (highly powerless)}} \\
\midrule

\vspace{0pt}Ch.\ 2 -- \emph{Roast Mutton} &
\vspace{0pt}\emph{yer, never, time, good, cook, none, mutton, shut, lots, yerself} &
\vspace{0pt}\includegraphics[width=0.95\linewidth]{figures/wc_ch02_roast_mutton} &
\vspace{0pt}This early encounter highlights the group’s inexperience and vulnerability. 
Frequent negation-related terms such as \emph{never}, \emph{none}, and \emph{shut} signal a 
lack of agency, while expressions like \emph{yerself} and \emph{lots} suggest troll dominance 
and mockery. \\[0.4em]

\vspace{0pt}Ch.\ 5 -- \emph{Riddles in the Dark} &
\vspace{0pt}\emph{precious, gollum, lost, pocketses, guess, nassty, curse, goblinses, preciouss, baggins} &
\vspace{0pt}\includegraphics[width=0.95\linewidth]{figures/wc_ch05_riddles_dark} &
\vspace{0pt}Bilbo’s riddling with Gollum registers as low dominance. Although he ultimately wins, his speech is tentative and reactive, motivated by fear of discovery, while Gollum’s distorted language and repeated \emph{precious} dominate the dialogue space. \\[0.4em]

\vspace{0pt}Ch.\ 6 -- \emph{Out of the Frying-Pan into the Fire} &
\vspace{0pt}\emph{goblins, dark, time, bit, mountains, burglar, quietly, baggins, great, miles} &
\vspace{0pt}\includegraphics[width=0.95\linewidth]{figures/wc_ch06_frying_pan} &
\vspace{0pt}Here the company remains mostly reactive, focused on survival rather than control. Frequent \emph{dark} and \emph{quietly} suggest concealment and caution; dialogue coordinates movement under pressure rather than asserting power, reflecting low dominance. \\

\end{longtable}
}%

% ====================================================
\section{Discussion}

The VAD line graph shows moderate valence, low arousal, and a steady rise in dominance --- an emotional arc defined more by composure than by extremes. This pattern aligns with Tolkien’s own description of \emph{The Hobbit} as “light-hearted” compared with the “more adult” and “more terrifying” \emph{The Lord of the Rings} \citep{tolkien1981}. The VAD trends reinforce this distinction, as emotional lows in \emph{The Hobbit} are brief and consistently counterbalanced by humour, wonder, or companionship.

The table provides additional nuance by showing how emotional tone is expressed through specific patterns of high-frequency words. Chapters with high valence --- such as “A Short Rest’’ and “A Warm Welcome’’ --- feature lexical fields dominated by communal, ceremonial, or restorative vocabulary (e.g., \emph{elves}, \emph{good}, \emph{king}, \emph{merry}). Conversely, low-valence chapters foreground words associated with confinement, ambiguity, or threat (e.g., \emph{lost}, \emph{dark}, \emph{curse}, \emph{goblins}). High-arousal chapters emphasize movement and danger (\emph{struck}, \emph{quicker}, \emph{dragon}), whereas low-arousal chapters rely on stillness and domesticity (\emph{sitting}, \emph{thinking}, \emph{mutton}). Shifts in dominance are likewise reflected through authoritative titles (\emph{king}, \emph{master}, \emph{friends}) or markers of vulnerability (\emph{none}, \emph{shut}, \emph{lost}). Taken together, the VAD trajectories and lexical evidence show that Tolkien’s emotional contour is not only quantitative but linguistically embodied: the dialogues feel the way they do because of the words characters repeatedly choose.

More broadly, the results demonstrate how computational methods can extend close reading. As \citet{elkins2025} argues, sentiment analysis can surface latent emotional structures that shape narrative experience. In \emph{The Hobbit}, those structures appear as alternating phases of tension and repose --- a rhythm that sustains engagement through variation rather than abrupt extremes. Through this lens, quantitative modelling does not replace interpretation but deepens it: mapping valence, arousal, and dominance reveals a narrative shaped by resilience and hope, even in moments of danger.

% ====================================================
\section{Conclusion}

Through a combination of computational analysis and literary interpretation, this study mapped the emotional contours of \emph{The Hobbit}'s dialogue using RegEx, the NRC-VAD lexicon, and Python-based visualization. Conceived within the framework of digital philology, it demonstrates how computational tools can extend traditional textual analysis by uncovering subtle emotional and linguistic patterns.

Analysis of chapter-level word frequencies further reinforces this structure: chapters marked by warmth or authority prominently feature communal or ceremonial vocabulary, while low-valence or low-dominance chapters foreground terms associated with threat, confinement, or hesitation. Together, these quantitative and lexical signals clarify how emotional tone is woven directly into the language characters use.

The analysis shows that Tolkien’s dialogue maintains a steady emotional equilibrium throughout the novel --- moderately positive in valence, generally low in arousal, and gradually increasing in dominance. This pattern reflects the tonal rhythm observed in the dialogue, where moments of tension and danger are followed by calm or humour, and instances of helplessness gradually give way to confidence. By transforming Tolkien’s language into measurable emotional data, the study makes visible tendencies often discussed qualitatively. The alternation of light and dark moods, or the recurring shift from fear to reassurance, can now be articulated in quantifiable terms that illuminate how \emph{The Hobbit} sustains its emotional balance.

In the end, visualizing Tolkien’s dialogue in terms of valence, arousal, and dominance provides a clearer view of the story’s emotional rhythm. By examining how the dialogue moves between fear and comfort, peril and peace, this analysis highlights the steady optimism that underpins \emph{The Hobbit}'s enduring appeal.

% ====================================================
\section{Limitations and Future Work}

While lexicon-based sentiment analysis provides a transparent framework, it has limits in accounting for contextual nuance. For instance, the word \emph{precious} (valence $\approx 0.83$ in the NRC-VAD lexicon) appears frequently in Chapter 5, ``Riddles in the Dark.'' Yet in Gollum's dialogue it carries connotations of menace and obsession rather than affection. This example shows how lexicon scores---though useful for identifying broad tendencies---cannot always capture the contextual or ironic meanings of specific words. Similarly, proper names such as \emph{Smaug} or \emph{Baggins} evoke emotional associations not reflected in static lexicons.

Future research could build on these findings by incorporating context-aware sentiment models that adjust meaning based on surrounding words, or by combining lexicon-based methods with machine learning approaches trained on Tolkien's work. Another valuable direction would be to compare dialogue and narration sentiment within \emph{The Hobbit}, following the general principle outlined by \citet{vishnubhotla2024}, to explore how emotional tone differs between the narrator's voice and the characters' speech. Comparative studies could also contrast \emph{The Hobbit} with \emph{The Lord of the Rings} to trace shifts in emotional pacing and maturity, or examine translations and adaptations to see how tone changes across languages and media.

% ====================================================
\section*{Appendix}

Because certain inputs and outputs (such as the source text) cannot be included here for copyright reasons, only permitted materials are provided. All project code and the available outputs can be found at the following link:
\begin{center}
\url{https://github.com/lilin6ykx/hobbit-sentiment-analysis.git}
\end{center}

% ====================================================

\end{document}